\documentclass{article}
\usepackage{spconf}
\usepackage{amsmath,graphicx,bm,booktabs}
\usepackage{color}

\title{Feature Fusion Encoder Decoder Network for Automatic Liver Lesion Segmentation}
\name{Xueying~Chen$^1$, Rong~Zhang$^1$, Pingkun~Yan$^2$}
\address{$^1$Department of Electronic Engineering and Information Science,\\
University of Science and Technology of China, Hefei,  China\\
$^2$Department of Biomedical Engineering,\\
Rensselaer Polytechnic Institute, Troy, NY 12180, USA
}
\begin{document}
\ninept
\maketitle
\begin{abstract}
Liver lesion segmentation is a difficult yet critical task for medical image analysis.  Recently, deep learning based image segmentation methods have achieved promising performance, which can be divided into three categories: 2D, 2.5D and 3D, based on the dimensionality of the models.  However, 2.5D and 3D methods can have very high complexity and 2D methods may not perform satisfactorily.  To obtain competitive performance with low complexity, in this paper, we propose a Feature-fusion Encoder-Decoder Network (FED-Net) based 2D segmentation model to tackle the challenging problem of liver lesion segmentation from CT images.  Our feature fusion method is based on the attention mechanism, which fuses high-level features carrying semantic information with low-level features having image details.  Additionally, to compensate for the information loss during the upsampling process, a dense upsampling convolution and a residual convolutional structure are proposed.  We tested our method on the dataset of MICCAI 2017 Liver Tumor Segmentation (LiTS) Challenge and achieved competitive results compared with other state-of-the-art methods.
\end{abstract}
\begin{keywords}
Liver lesion segmentation, deep learning, encoder-decoder, attention, feature fusion.
\end{keywords}

\section{Introduction}
\label{sec:intro}

Liver cancer is one of the most commonly seen cancers with high mortality rate in the modern world.  Computed tomography (CT) is often used for tumor diagnosis and treatment.  Manual liver lesion segmentation on CT is time-consuming, difficult to reproduce, and easily influenced by personal subjective experience.  Therefore, fully automatic method for liver lesion segmentation needs to be developed.
However, automatic liver lesion segmentation is a very challenging task due to the following reasons.  First, CT images often have low contrast for soft tissues and the imaging noise makes it difficult to segment the lesions.  Second, the position and the shape of the lesion vary significantly.  Last but not the least, the lesion boundaries are unclear and the sizes of most lesions are relatively small.

To solve the problems mentioned above, a number of methods based on deep learning have been proposed for the liver lesion segmentation task in the past several years.  Generally, these learning methods can be divided into three categories: 1) 2D models, such as Cascaded-FCN \cite{Christ2016Automatic}, FCN based on VGG-16 \cite{Ben2016Fully}; 2) 2.5D models, such as U-Net using residual connections \cite{han2017automatic} and method exploiting inter-slice similarity \cite{zhu_exploiting_2018}; 3) 3D models, such as Densely connected volumetric ConvNets \cite{zhu_ijcnn}, H-DenseUNet \cite{Li2017H}, 3D FCN \cite{dou20163d}.  Even though the previous methods have obtained much improved performance, there still exist a lot of potentials.  For example, previous 2D networks cannot fully utilize the depth information of the CT image and thus its segmentation accuracy is low.  On the other hand, 2.5D and 3D networks are computationally expensive and have very high requirement for hardware configurations.  Given the same computing resources, the large amount of parameters and heavy computation limit the depth and complexity of the 3D networks.

With the aforementioned considerations, we propose a Feature-fusion Encoder-Decoder Network (FED-Net) based 2D model for image segmentation.  First, we design a novel feature fusion method based on the attention mechanism, which can effectively embed more semantic information into low-level features and improve the current feature fusion mode based on the U-Net architecture network.  Then, to compensate for the information loss during the upsampling process, we propose a dense upsampling convolution instead of using the traditional upsampling operations and also add residual convolutional blocks to refine the rough boundary of the target.

\begin{figure*}[tb]
     \begin{minipage}{0.5\linewidth}
            \centering 
	    \includegraphics[width=3.5in]{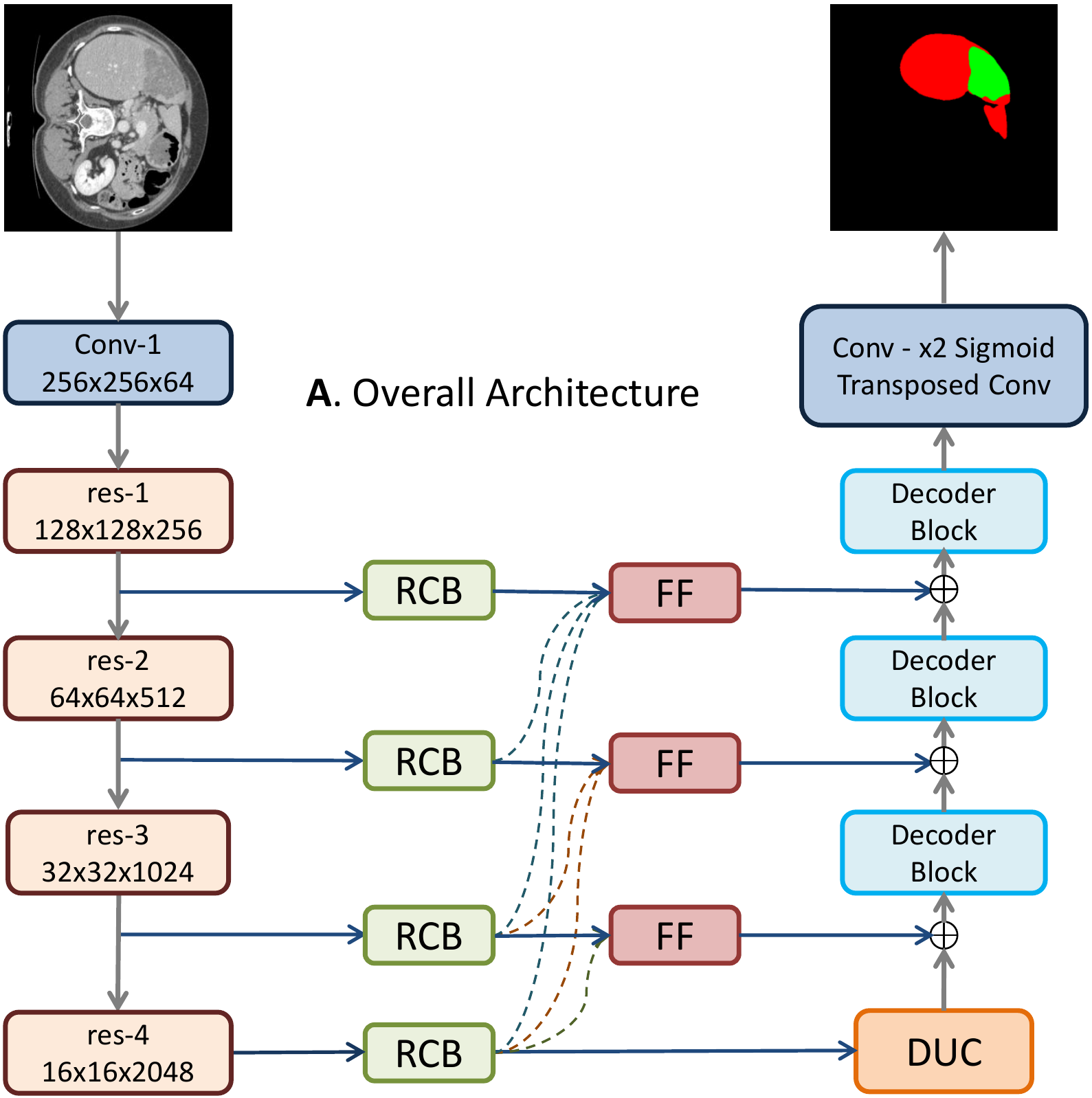} 
     \end{minipage}
     \hfill 
     \begin{minipage}{0.5\linewidth}
           \centering 
     	   \includegraphics[width=3.4in]{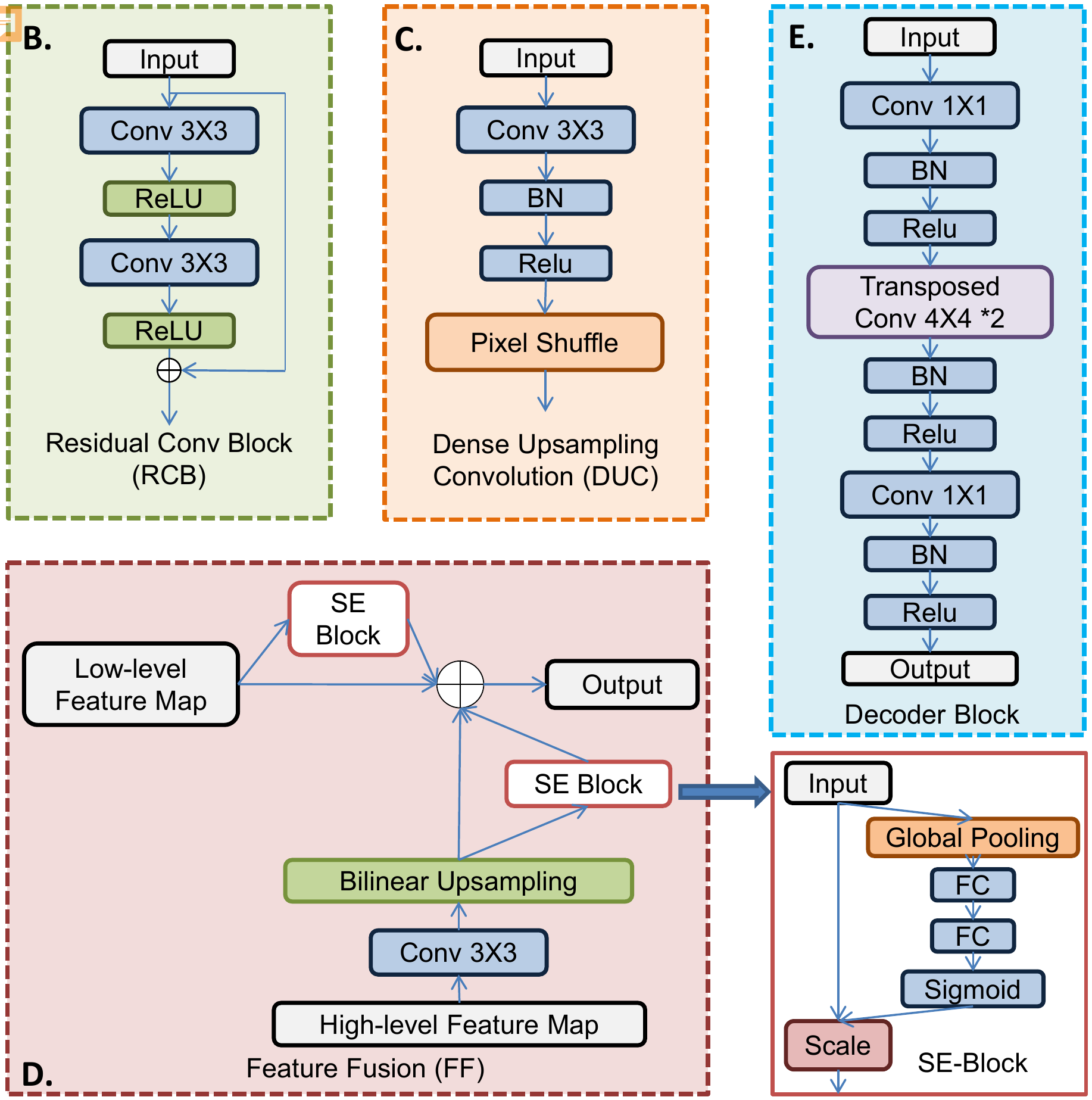}
     \end{minipage}
\caption{The overall architecture of our proposed approach and the individual components of the FED-Net.  The encoder part uses ResNet-50 with the fully connected layer removed.  \textbf{RCB} - Residual Convolution Block, \textbf{FF} - Feature Fusion Based on Attention, \textbf{DUC} - Dense Upsampling Convolution.  \textbf{Decoder block} - first reduce the number of filters by 1$\times$1 convolution, then perform transposed convolution, and finally restore the number of filters by 1$\times$1 convolution.}
\label{fig:network}
\end{figure*}

\section{METHOD}
\label{sec:method}

Among the many aforementioned problems of convolutional neural network (CNN) based image segmentation methods, one major issue with semantic segmentation is due to the pooling layers.  Pooling layer not only expands the receptive field but also aggregates the context while discarding the location information. However, semantic segmentation requires exact alignment of class maps and thus, needs the location information to be preserved.  To solve such problem, many encoder-decoder architectures have been proposed.  In the encoder-decoder framework, the encoder gradually reduces the spatial dimension of the feature maps and the decoder gradually recovers the object details and spatial dimensions.  Skip connections are usually used between the encoder and decoder, which help the decoder to better retain the details of the target, especially for situations with small sample sizes.  The widely adopted U-Net architecture \cite{ronneberger2015u} is a typical encoder-decoder structure.  There are quite a few models that fuse features in this fashion, for example RefineNet \cite{Lin2017RefineNet} and ExFuse \cite{zhang2018exfuse}.  However, typical U-Net based networks directly fuse high-resolution features from encoder path to the upsampled decoder output through either addition, multiplication, or concatenation, which may not be able to efficiently use features from different levels.  Our work suggests that all levels of features should contribute more directly to semantic segmentation and the importance of features should be weighed differently.

In this paper, we propose a new method for feature fusion with an attention mechanism.  To refine the rough boundary due to upsampling operation, residual convolution blocks (RCBs) are inserted before feature fusion.  Then after the feature fusion, in order to make up for the lost location information during the upsampling process, the dense upsampling convolution (DUC) is applied.  An overview of the proposed method is given in the left side of Fig.~\ref{fig:network}.

\subsection{FED-Net}
\label{ssec:refine}

In this subsection, details of the proposed Feature-fusion Encoder-Decoder Network (FED-net) for medical image segmentation are provided.  The encoder part use pre-trained model on the ImageNet - ResNet50 \cite{he2016deep}.  We divide the pre-trained ResNet-50 into four blocks according to the resolution of the feature map, and add the residual convolution block after the four blocks separately.  Feature fusion is then performed on the output of each block, and finally each unit is sent to the decoder part.

\textbf{Attention based Feature Fusion.}  For convolutional neural networks, the low-level general extracts the low-level features and contains little semantic information, which has been confirmed by ZF-Net \cite{zeiler2014visualizing}.  As mentioned above, the typical U-Net based networks directly fuse high-resolution features from encoder path to the upsampled decoder output, which may not be able to efficiently use features from different levels.  Inspired by SE-NET \cite{hu2017squeeze}, we proposed a feature fusion method based on attention mechanism to fuse features of different levels.  We generalize the fusion method into the form of
\begin{align}
H_{l} = SE(x_{l}) + \sum_{i=l+1}^L SE((U(x_{i}))
\label{eqn:AFF}
\end{align}
where $L$ is the number of feature levels.  The feature fusion method is illustrated in Fig.~\ref{fig:network}-D.  Our feature fusion first extracts multi-resolution features from all levels of the encoder.  Then the feature fusion module at each level fuses the features at its current level and also the levels above it.  Where the function $U(\cdot)$ - upsampling operation is performed on the high-level features, and the SE-Net is executed on all the features that need to be fused.  The detail of SE-block are shown in the lower right corner of Fig.~\ref{fig:network}.
With this feature fusion approach, more useful features can be extracted by assigning different weights to the features from different based on their contribution to the final segmentation results.  Then, the importance of those features which have more impact on the segmentation results can be emphasized.

\textbf{Residual Convolution Block.}  A residual convolution block (RCB) that does not include batch normalization is added after the four different resolution blocks that are extracted from the encoder part.  The RCB is easy to train using the ``shortcut connection'' method and further extracts intermediate features to improve the results.  The details of the RCB structure is shown in Fig.~\ref{fig:network}-B.

\begin{figure}[tb]
\includegraphics[width=\columnwidth]{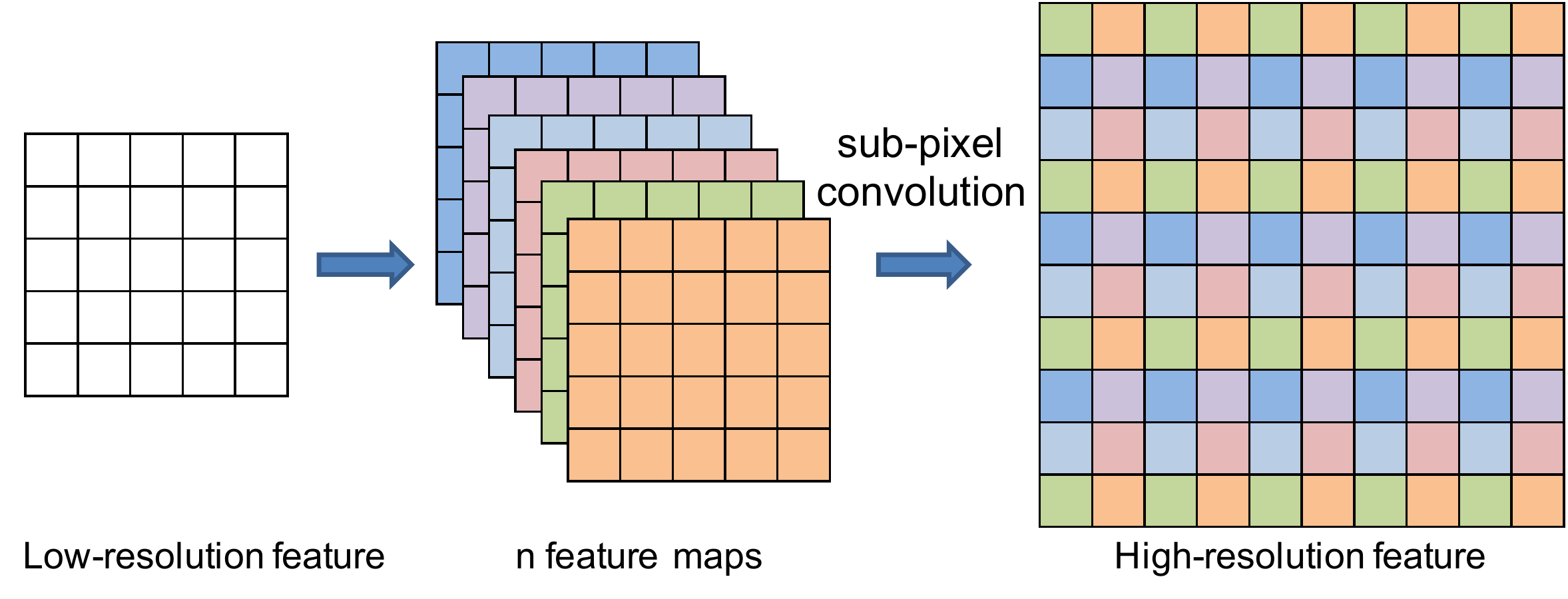}
\caption{Sub-pixel convolutional module with pixel shuffling.}
\label{fig:duc}
\end{figure}

\textbf{Dense Upsampling Convolution.}  To compensate for the details lost during the upsampling process, dense upsampling convolution (DUC) blocks are used which is a sub-pixel convolution with pixel shuffling \cite{Shi2016Real}. The structure of DUC is in Fig.~\ref{fig:network}-C.  The details of pixel shuffling are shown in Fig.~\ref{fig:duc}, which is mainly to periodically merge multiple low-resolution feature maps into a high-resolution map according to specific positions determined by a ``period shuffling'' operation.  We use DUC as the first block of the decoder part. That is to have the highest level feature map not only contain highly abstract features but also stay in relatively higher resolution.  This is designed to deal with the commonly seen block artifact from fully convolutional neural networks based segmentation methods.

\subsection{Loss Function}
\label{ssec:loss}

Considering the imbalance between the positive and negative samples of the data set, we combined the binary Cross Entropy with the Jaccard index.  The overall loss of our proposed method is given below:
%
\begin{align}
L(y, \hat{y}) = & (\omega_{1}-1)*y \log\hat y-\omega_{1}*(1-y)\log(1-\hat y) \nonumber \\
  & -\omega_{2}*log( \frac{\vert y \cap  \hat y \vert + \epsilon}{\vert y \vert + \vert  \hat y \vert - \vert y \cap  \hat y \vert + \epsilon})
\label{eqn:loss}
\end{align}
%
where $y$ represents the ground truth segmentation of a slice and $\hat{y}$ denotes the predicted mask of the slice. 

The first two terms on the right of Eqn.~(\ref{eqn:loss}) define a weighted binary cross entropy.  The weighting parameter $\omega_1$ can help alleviate the problem of data imbalance between positive and negative samples.  In our application, the volume of lesion is quite small compared to the rest of the imaging volume.  The third term of Eqn.~(\ref{eqn:loss}) contains the generalized Jaccard index that is commonly used to measure the quality of segmentation result by computing the similarity between two segmentation as the size of the intersection divided by the size of the union.  The generalized Jaccard index is used for two reasons.  On the one hand, the index is differentiable, where back propagation can be conviently computed.  On the other hand, it is a proper distance metric relative to the Dice index.  In the third term on the right side of the above equation, $\epsilon$ is a very small number to prevent the denominator being zero.  In our work, we used $\epsilon = 1\times 10^{-15}$.

\subsection{Hierarchical Segmentation}

For liver tumor segmentation, we take a two-step process.  The first step is to segment the liver with a baseline segmentation network and the second step is to segment the lesion with the FED-Net.  Our baseline network is a ``U-Net'' network using the ResNet-50 \cite{he2016deep} as the encoder part. 

In the first stage of the liver segmentation, we used the entire volume for training and validation.  In the second stage for lesion segmentation, in order to reduce the number of negative samples and the amount of computation,  we sent a slice containing only the liver into the network. In the test phase, only the liver slice obtained from the first step segmentation were used.  To alleviating the imbalance between the positive and negative samples of the dataset, we implemented a probability extraction strategy to extract positive and negative samples from the dataset during training.  The positive samples were extracted with a probability of 0.9, and the negative samples were extracted with a probability of 0.1.

To make full use of the third dimension information of the CT image and without adding extra calculations, we sent the adjacent three slices as the three channels of the intermediate slice into the network.

\section{EXPERIMENTS AND RESULTS}
\label{sec:result}

\subsection{Dataset and Preprocessing}
\label{sec:dataset}

We used the publicly available dataset of MICCAI 2017 LiTS Challenge as the training of our proposed method.  The LiTS datasets contains 131 and 70 contrast-enhanced 3D abdominal CT scans for training and testing, respectively.  We tested our method on the 70 LiTS challenge test cases.  The dataset was acquired from a number of different clinical sites with different scanners and protocols.  Each slice in all CT scans has a fixed size of 512$\times$512 pixels with the number of slices per scan varies from 42 to 1026.  The image resolutions are also different from scan to scan.

In order to remove irrelevant information about other organs and tissues in the CT scans for liver lesion segmentation, we cut the the image intensity values of all CT scans to the range of [-200, 250] HU.  After all CT scan HU values were truncated, we normalized all slice intensities into the range [0,1] with min-max normalization.

\subsection{Implementation Details}
\label{sec:implement}

Our proposed FED-Net was implemented using the pytorch package.  Due to its efficiency, the experiment used a single NVIDIA GTX 1080 GPU with 8 GB memory.  We used stochastic gradient descent method for optimization with momentum of 0.9 and weight decay of 0.0001.  We used the CT scans from 0 to 119 and 120 to 130 of the LiTS dataset as our training set and validation set, respectively.  Simple data augmentation including horizontal flipping and vertical flipping were performed.

In the post-processing stage: (1) The threshold for the output of liver segmentation is 0.5, and 3D connect-component labeling was used.  (2) The threshold for the output of lesion segmentation is 0.3.  On the basis of (1), a bounding-box was taken for the result of liver segmentation, and then the liver segmentation result with bounding-box was merged with the lesion segmentation result to obtain the final lesion segmentation result.

\subsection{Ablation Analysis of FED-Net}
\label{sec:results}

In this subsection, we will make step-by-step comparisons to evaluate our approaches proposed in Section~\ref{sec:method}.
In Section~\ref{sec:method}, we propose Attention based Feature Fusion(FF) to effectively assign weights to embed more semantic information into low-level features.  To compensate for the missing details during the upsampling process, we propose Dense Upsampling Convolution(DUC) and Residual Convolutional Block (RCB).  To demonstrate the effectiveness of our method, we tested the performance of each component separatively.  The results are presented in Table~\ref{tab:ablation}.  From the results, we can find that the performance has increased correspondingly with the addition of FF, DUC and RCB.  When we used the feature fusion method without SE-Block, we found that the performance hardly improve, and when we used our feature fusion method with SE-Block, the result increase by 0.8\%.

\begin{table}[ht]
\caption{Liver lesion segmentation Dice scores of our method on the LiTS challenge test cases under different ablations.}
\label{tab:ablation}
\centering
\begin{tabular}{c|cc}
\toprule
Model& Per case& Global\\
\midrule
Baseline& 0.635& 0.745\\
Baseline + RCB& 0.644& 0.776\\
Baseline + FF& 0.636& 0.756\\
Baseline + FF with SE-Block& 0.643& 0.775\\
Baseline + DUC& 0.644& 0.772\\
Baseline + RCB + FF + DUC & \textbf{0.650} & \textbf{0.766}\\
\bottomrule
\end{tabular}
\end{table}

\begin{table}[ht]
\caption{Leader board Dice scores of 2017 Liver Tumor Segmentation (LiTS) Challenge}
\label{tab:Leaderboard}
\centering
\begin{tabular}{l|cc}
\toprule
Method  & Per case & Global\\
\midrule
H-DenseUNet \cite{Li2017H}& 0.722& 0.824\\
Res-UNet \cite{han2017automatic}& 0.670& -\\
Cascaded ResNet \cite{bi2017automatic}& 0.640& -\\
3D AH-Net  \cite{liu20183d}& 0.634& 0.834\\
FCN+Detector \cite{bellver2017detection}& 0.590& -\\
\textbf{ours}& 0.650 & 0.766\\
\bottomrule
\end{tabular}
\end{table}

We compare our results to a number of excellent methods, Table~\ref{tab:Leaderboard} shows the details.  Most of the first few methods are based on 2.5D or 3D models, among them, H-DenseUNet and 3D AH-Net are 3D models, Res-UNet and FCN+Detector are 2.5D models, and Cascaded ResNet is 2D models, our model shows very promising performance.  It isn't too far away from other leading 3D and 2.5D methods, given it is in 2D and the number of parameters is much less.  It is light weighted and can be easily deployed.  We hope to further improve the performance in future by extending it to 3D and also with other enhancements.

\section{CONCLUSIONS}
\label{sec:conclusion}

In this work, we proposed a Feature-fusion Encoder-Decoder Network (FED-Net) based 2D deep learning model for liver lesion segmentation.  Our feature fusion method uses the attention mechanism.  By assigning different weights to different features according to the contribution of the feature to the final segmentation result, more useful features can be extracted.  Our feature fusion method can effectively embed more semantic information into low-level features.  We also propose a dense upsampling convolution instead of some conventional upsampling operations to compensate for the missing details during the upsampling process and add residual convolutional blocks to refine the target boundary information.  Our approach demonstrates very promising performance relative to 2.5D and 3D models, and our 2D model is lightweight and can be easily deployed.  In addition, our network has good generalization capabilities and can be applied to other areas of segmentation tasks, not just medical images.

\bibliographystyle{IEEEbib}
\bibliography{refs}

\end{document}